\documentclass[a4paper,conference]{IEEEtran}
\IEEEoverridecommandlockouts
\usepackage{cite}
\usepackage{amsmath,amssymb,amsfonts}
\usepackage{algorithmic}
\usepackage{graphicx}
\usepackage{multirow}
\usepackage{textcomp}
\usepackage{xcolor}
\usepackage{adjustbox}
\def\BibTeX{{\rm B\kern-.05em{\sc i\kern-.025em b}\kern-.08em
    T\kern-.1667em\lower.7ex\hbox{E}\kern-.125emX}}
\begin{document}

\title{Combining Similarity and Adversarial Learning to Generate Visual Explanation: \\ Application to Medical Image Classification 
}

\author{\IEEEauthorblockN{Martin Charachon}
\IEEEauthorblockA{\textit{Incepto Medical - Université Paris-Saclay, CentraleSupélec, MICS}}
\and
\IEEEauthorblockN{Céline Hudelot}
\IEEEauthorblockA{\textit{Université Paris-Saclay, CentraleSupélec, MICS}}
\and
\IEEEauthorblockN{Paul-Henry Cournède}
\IEEEauthorblockA{\textit{Université Paris-Saclay, CentraleSupélec, MICS}}
\and
\IEEEauthorblockN{Camille Ruppli}
\IEEEauthorblockA{\textit{Incepto Medical}}
\and
\IEEEauthorblockN{Roberto Ardon}
\IEEEauthorblockA{\textit{Incepto Medical}}
}

\maketitle

\begin{abstract}
Explaining decisions of black-box classifiers is paramount in sensitive domains such as medical imaging since clinicians confidence is necessary for adoption. 
Various explanation approaches have been proposed, among which perturbation based approaches are very promising. Within
this class of methods, we leverage a learning framework to produce our visual explanations method.
From a given classifier, we train two generators to produce from an input image the so called similar and adversarial images. The similar image shall be classified as the input image whereas the adversarial shall not. Visual explanation is built as the difference between these two generated images. Using metrics from the literature, our method outperforms state-of-the-art approaches. The proposed approach is model-agnostic and has a low computation burden at prediction time. Thus, it is adapted for real-time systems. Finally, we show that random geometric augmentations applied to
the original image play a regularization role that improves several previously proposed explanation methods. We validate our approach on a large chest X-ray database.
\end{abstract}


\section{Introduction}\label{sec:introduction}
Deep neural models have enabled to reach high performances on various applications and in particular in medical image analysis\cite{skincancerEsteva2017}.
In this field, additionally to high accuracy, it is critical to provide interpretative explanations of the system decision since the clinicians' confidence in the system is at stake\cite{Holzinger2017WhatDW}.
Thus, several methods were proposed to address the visual explanation problem, ranging from saliency maps\cite{simonyan_deep_2014, smilkov_smoothgrad:_2017, hechtlinger_interpretation_2016, sundararajan_axiomatic_2017} and class activation mapping methods\cite{zhou_learning_2016,
selvaraju_grad-cam:_2017} to perturbation maps\cite{fong_interpretable_2017, dabkowski_real_2017}. However the problem is still open since there is no general consensus on their performances.
Independently, under the motivation of model safety, several methods were proposed to generate "fake"
images that "fool" classification algorithms.
Such "fooling" images are generated either by the addition of a small perturbation
on top of the input image\cite{goodfellow_explaining_2014,  Elliott2019AdversarialPO, xiao_generating_2019}, or as a complete new image, very close to the input\cite{zhang_generating_2019}. Most of these works point out that adversarial generation allows to study the model robustness and fragility but very few make links with the explainability problem.

In this work, we propose to leverage adversarial generation methods to produce interpretative
explanation of classifier's decision.
Inspired by \cite{dabkowski_real_2017} and  \cite{zhang_generating_2019} who both train model to generate respectively explanation masks and adversaries, we propose to train a model to generate images that capture discriminative structures with the following key contributions:
\begin{itemize}
    \item A new framework of explanation is proposed.
    We define the explanation of a classifier's decision as the difference between a regularized adversarial example and
    the "projection" of the original image into the space of adversarial generated samples.
    \item We introduce a new optimization workflow that combines the training of an
    adversarial generator and of a similar generator that "projects" the original image into the adversarial space.
    \item We propose a new method that greatly improves the use of several explanation
    methods as means to localize decisive objects in the image: Namely, averaging of
    registered explanation results built upon random geometric augmentations of the input image.
\end{itemize}
We validate our algorithm on a binary classification problem (pathological/healthy) of a large database of X-ray chest images.
We compare our technique to state of the art approaches such as Gradient\cite{simonyan_deep_2014}, GradCAM\cite{selvaraju_grad-cam:_2017}, BBMP\cite{fong_interpretable_2017}, Mask Generator\cite{dabkowski_real_2017}.
\section{Related Work}\label{sec:related_work}
Explaining classifiers decisions via some visual map has been the subject of several contributions. We here classify in three categories a selected few.

\subsection{Back-propagation-based methods}\label{subsec:back-propagation-based-methods}
These methods leverage, for neural networks and for a given image, back-propagation of small variations of the model's prediction\cite{simonyan_deep_2014}.
While providing interesting results, these methods tend to produce noisy explanation maps since any variations of the model's output is considered.
Many contributions are thus focused on building sharper and smoother explanation maps\cite{zeiler_visualizing_2013, springenberg_striving_2014, smilkov_smoothgrad:_2017, sundararajan_axiomatic_2017}.
In\cite{zhou_learning_2016}, explanation maps are produced by upsampling activations from the last convolutional layer to the input image size, GradCAM\cite{selvaraju_grad-cam:_2017} (or its application in medical
domain\cite{rajpurkar_chexnet:_2017}), builds on this work by computing the gradient of the output with respect to the last convolutional layer (and not only with respect to the model's prediction) generating compelling results. For an exhaustive review, the reader can refer to \cite{adebayo_sanity_2018} and \cite{Rebuffi2020ThereAB}.
As a limitation, these approaches are not model-agnostic (limited to neural networks) and need access to intermediate layers.

\subsection{Iterative perturbation-based methods}\label{subsec:iterative-perturbation-based-methods}
The principle of this approaches is to exploit the effects of perturbations to the input image
on the model's output\cite{zeiler_visualizing_2013}.
For instance, LIME\cite{ribeiro_why_2016} proposes a local explanation by perturbing random segments of the input image and training
a linear classifier to predict the importance of each segment for the classifier's prediction.\\
The authors of\cite{fong_interpretable_2017} take this idea further by defining their explanation maps as the result of an optimization procedure over the input image and the model to explain.
Considering a fixed perturbation, their approach consists in learning the maps that maximally impact
the model or on the contrary the maps that enable to preserve its performance. Similarly, building on \cite{chang_explaining_2019} for the medical imaging domain,\cite{uzunova_interpretable_2019, major_interpreting_2020} adopt a similar optimization formulation but focus on the
perturbations.
They use generative methods to, respectively, perturb pathological images by local in-painting
healthy tissues within pathological images or completely reconstructing a healthy image.\\
As noisy outputs is a major concern within these methods, some authors focus on regularization terms~\cite{Fong2019UnderstandingDN} and others
filter gradients during back-propagation~\cite{Wagner2019InterpretableAF}. Different optimization formulations were also
introduced\cite{dhurandhar_explanations_2018, hsieh2020evaluations}.
In\cite{khakzar_explaining_2019}, an explanation is generated through features perturbation at different levels of the neural network.
In\cite{woods_adversarial_2019, Elliott2019AdversarialPO} the optimization problem is revisited as an adversarial example generation, where the adversarial perturbation is sought in a
regularized  and restricted space.\\
Note that all these methods have in common the necessity to solve, for each image, an optimization problem in order to produce an explanation map.
This translates into a high computational cost, as several iterations are needed for convergence (often inappropriate when a real time response is expected).
Moreover, since the optimization problem is solved for every image, over-fitting issues arise.
Explanation maps often contain features not linked to the models' behaviour but only to the image
being processed.
Strong and elaborated regularization is thus a necessity.

\subsection{Trained perturbation-based methods}\label{subsec:generative-perturbation-based-methods}
In order to alleviate computational needs of iterative perturbation-based methods, \cite{dabkowski_real_2017}
proposes to evolve to an optimization problem on the whole database, thus learning a masking model.
In medical imaging,\cite{DBLP:journals/corr/abs-1807-07784} uses the same approach on a single class problem.
Despite the benefits of this optimization strategy, two main drawbacks remain.
First, perturbations are provided as parameters to be set and adapted manually.
Their choice is impacted not only by the database and the considered classifier but also by the training of this classifier (e.g. a random
noise perturbation has no effects on models trained to be robust to this noise).
Since perturbations are manually selected, residual adversarial artefacts (without any link to the explanation) are still generated.
Second, a costly hyperparameters tuning is needed to control the size of the generated explanation masks.


\section{Methodology}\label{sec:methodology}
We present our methodology to generate explanations for image classification outcomes.
As for the methods exposed in section~\ref{sec:related_work}, our explanation is given as a visual explanation map where higher values code for more important areas on the image w.r.t to the classifier decision.
For the sake of simplicity, we present the rationale behind our approach in the case of a binary classification problem, the extension to the multi-class case being presented in section~\ref{subsec:multi-class}.
Let $f_c$ be the studied classifier outputting a classification score in the range $[0,1]$ . Without loss of generality, we assign label 1 (resp. label 0) to an input image if its classification score ($f_c$ value) is over 0.5 (resp. under 0.5). In the case of a different threshold one can apply for instance a piece-wise linear transformation to $f_c$ to satisfy this condition.


\subsection{Explanation through similar and adversarial generations}\label{subsec:method}

\subsubsection{Adversarial naive formulation}

A naive, yet novel, approach to reach our objective is to combine a trainable masking model \cite{dabkowski_real_2017} with adversarial perturbations for visual explanations\cite{Elliott2019AdversarialPO}.
The visual explanation map is then given by the
difference between the input image and its adversary. This method is no longer dependent on the choice of a perturbation function since the adversarial sample "learns" this perturbation. 
For an input image $x$ we define the \textbf{\textit{naive}} explanation as
\begin{equation}\label{eq:n_explanation}
	E^{naive}_{f_c}(x) = |x - \bar{g}_{a}(x) |
\end{equation}
where $\bar{g}_{a}$ is a model obtained via a training process with the goal of "fooling" the classifier $f_c$ while producing an image "very close" to $x$, written as follows:
\begin{equation}\label{eq:n_ga}
\begin{array}{lccl}
     \bar{g}_{a} &=&
     \hspace{-1cm} \underset{g_a}{\mathrm{argmin}}  &
     \hspace{-0.5cm} \mathbb{E}_x\left [  \left \|x - g_a(x)\right \|_2   \right ] \\
     && \hspace{-1cm} \scriptstyle s.t.\, f_c(g_a(x)))= 1-f_c(x) &
\end{array}
\end{equation}
The mean value is taken over a training data set.
Generating a visual explanation using~\eqref{eq:n_explanation} and~\eqref{eq:n_ga} effectively counterbalances drawbacks of \cite{dabkowski_real_2017} \& \cite{Elliott2019AdversarialPO} but despite the regularization expected from the learning process, visual explanations are often corrupted by noise, highlighting regions which clearly should have no impact on the classifiers decision (see section~\ref{subsec:weak-localization-evaluation}).

\subsubsection{Similar-Adversarial formulation}

Why does the \textbf{\textit{naive}} formulation generate incoherent visual explanations?
We argue that the flaw resides in the definition of explanation as expressed in equation ~\eqref{eq:n_explanation}.
Comparing the original image with its generated adversarial sample exposes the method to a risk of reconstruction error. Some details of the original image can be absent from the generated 
adversarial sample. However these details are not discriminative for the classifier in the sense that their sole presence would not 
change the classification score. 
More formally, the adversarial sample belongs to the target space of $\bar{g}_a$ ($\chi_a$) which is different from the space of original images ($\chi_o$).
The comparison between $x$ and $\bar{g}_a(x)$ inherits from the differences between $\chi_o$ and $\chi_a$ and these differences are not explicitly linked to the explanation problem by Equation~\eqref{eq:n_ga}. 
Since we do not have control on the original image space $\chi_o$, we propose to mitigate this
reconstruction risk by defining the explanation mask as the difference between the adversary
$\bar{g}_a(x)$ and the closest element to $x$ in $\chi_a$ on which $f_c$ returns the same value as
$x$. We call this element the \textit{similar} image and it is denoted by $\bar{g}_s(x)$. $\bar{g}_s$ is the function mapping images to their similar counterparts. The rationale is to reduce the reconstruction error so that $E_{f_c}(x)$ only contains values related to the classifiers' decision and reads
\begin{equation}\label{eq:explanation}
	E_{f_c}(x) =  |\bar{g}_{s}(x) - \bar{g}_{a}(x)|
\end{equation}
Denoting $\chi_s$ the target space of $\bar{g}_s$, both $\bar{g}_s$ and $\bar{g}_a$ are built through a joint optimization process aiming to make $\chi_s$ and $\chi_a$ as "close" as possible while satisfying $f_c(\bar{g}_s(x)) = f_c(x)$ and $f_c(\bar{g}_a(x)) = 1 - f_c(x)$:
\begin{equation}\label{eq:gs_ga}
\begin{array}{lccl}
     (\bar{g}_s, \bar{g}_a) \hspace{-0.25cm} &=&
      \hspace{-1cm} \underset{g_s, g_a}{\mathrm{argmin}}  & \hspace{-1cm}
     \mathbb{E}_x \left (
            \begin{array}{l}
             d_{\chi_o,\chi_s}(x, g_s(x)) \ + \\
             d_{\chi_o,\chi_a}(x, g_a(x)) \ + \\
             d_{\chi_s,\chi_a}(g_s(x), g_a(x))
             \end{array}
        \right ) + d(g_s,g_a)
             \\
     &&  \hspace{-1cm} \scriptstyle  s.t & \\
     &&  \hspace{-1cm} \scriptstyle f_c(g_s(x)))= f_c(x) & \\
     &&  \hspace{-1cm} \scriptstyle f_c(g_a(x)))= 1-f_c(x) &
\end{array}
\end{equation}
where $d_{\chi_o,\chi_s}$, $d_{\chi_o,\chi_a}$ and $d_{\chi_s,\chi_a}$ are distance functions between elements of the different image spaces while $d(g_{s}, g_{a})$ is a measure between the two functions. Henceforth, we refer to $\bar{g}_{s}$ and $\bar{g}_{a}$ as the similar and adversarial generators respectively.

\subsection{Weaker formulation: Objective function}\label{subsec:obj_func}
\begin{figure}[htbp]
\centering
  \includegraphics[width=.8\linewidth]{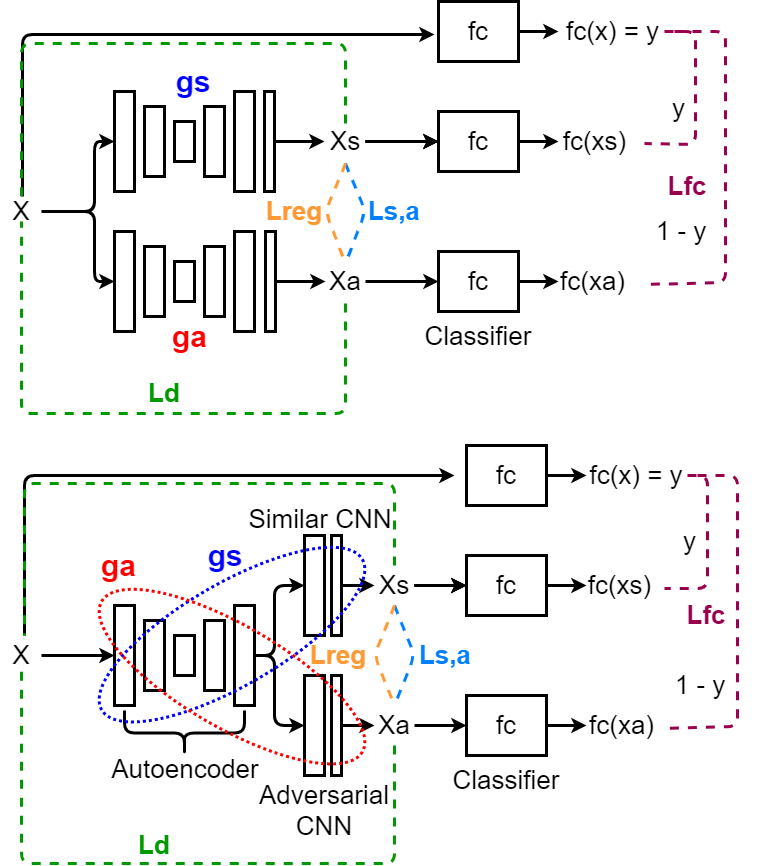}
  \caption{Overview of Duo $AE$ (Top) and Single $AE$ (Bottom)}
  \label{fig:duo_single_gen}
\end{figure}

We propose to solve a weak formulation of the previous constrained optimization problem ~\eqref{eq:gs_ga}.
We search for both similar and adversarial generators as minimizers of the following unconstrained problem
\begin{equation}\label{eq:loss}
 (\bar{g}_s, \bar{g}_a) = \underset{g_s, g_a}{\mathrm{argmin}} 
    \left\{ \hspace{-2mm}
    \begin{array}{ll}
        & \hspace{-0.5cm} \mathbb{E}_{x}
        \left(
        \begin{array}{ll}
            L_{d}(x, g_s(x), g_a(x)) & \hspace{-0.2cm} + \\
            L_{f_c}(x, g_{s}(x), g_{a}(x)) & \hspace{-0.2cm} + \\
            L_{reg}(x, g_{s}(x), g_{a}(x))  & \hspace{-0.2cm}
        \end{array}
        \right) \\\\  + &  L_{s,a}(g_{s}, g_{a}) 
    \end{array} \hspace{-3mm}
\right\}
\end{equation}
$L_d$ is a similarity loss that accounts for the term $d_{\chi_o,\chi_s} + d_{\chi_o,\chi_a} + d_{\chi_s,\chi_a}$ in equation~\eqref{eq:gs_ga} and enforces the 
proximity between $x$, $g_s(x)$ and $g_a(x)$. $L_{f_c}$, the classification loss, is a weak formulation of the classification constraints
in~\eqref{eq:gs_ga} enforcing the similarity between $f_c(x)$ and $f_c(g_s(x))$ and their dissimilarity 
with $f_c(g_a(x))$. $L_{s,a}$ enforces the similarity between $g_s$ and $g_a$ ($d(g_s, g_a)$). In addition to the terms of~\eqref{eq:gs_ga}, $L_{reg}$ acts on the
difference ($g_s(x) - g_a(x)$) to enforce regularity.
An embodiment of optimization problem \eqref{eq:loss} when using neural networks is given in Figure~\ref{fig:duo_single_gen} (see section \ref{subsec:generator_imple}).
We next specify the choices made in our method for each of the terms in Equation~\eqref{eq:loss}.

\subsubsection{Similarity Loss $L_d$}
Is defined as
\begin{equation}\label{eq:Ld}
L_{d}(x, g_s(x), g_a(x)) = 
          \begin{array}{l}
          \alpha_1 \left \|x-  g_s(x)\right \|_2 + \\
          \alpha_2 \left \|x -  g_a(x)\right \|_2 + \\
          \alpha_3 \left \|g_s(x) -  g_a(x)\right \|_2 + \\ 
          \alpha_4 \left \|g_s(x) -  g_a(x)\right \|_1 
          \end{array}
\end{equation}
where parameters $\alpha_1, \,\alpha_2, \,\alpha_3, \,\alpha_4 \in \mathbb{R}$ adjust the importance attached to the different terms. Combining $L_1$ and $L_2$ norms to enforce similarity between $g_s(x)$ and $g_a(x)$ produces better results experimentally (as in \cite{zhang_generating_2019}).
\subsubsection{Classification Loss $L_{f_c}$}
Is defined as
\begin{equation}\label{eq:Lfc}
L_{f_c}(x, g_s(x), g_a(x)) = 
    \begin{array}{l}
        \beta_1  L_{bce}(f_c(x), f_{c}(g_s(x)) + \\ \beta_2 L_{bce}(1-f_c(x), f_{c}(g_a(x))
    \end{array}
\end{equation}
where $\beta_1, \,\beta_2 \in \mathbb{R}$ are weighting parameters and $L_{bce}$ is the binary cross entropy loss.
This term accounts for the weak formulations of constraints in~\eqref{eq:gs_ga}, favoring classifier $f_c$ to act on $g_s(x)$ as it acts on $x$ and in the opposite manner for $g_a(x)$.
\subsubsection{Generator Loss $L_{s,a}$}
Is a measure of the distance between the two generators.
In the particular case (see section~\ref{subsec:generator_imple}) where they both are neural networks (parameterized by $w_s$ and $w_a$ respectively) we used the following metric
\begin{equation}\label{eq:gw}
L_{s,a}(g_s(., w_s), g_a(.,w_a)) = \gamma \left \| \sum_k w_s^k -  w_a^k\right \|_2
\end{equation}
where we assume generators $g_s$ and $g_a$ to have the same architecture. $\gamma \in \mathbb{R}$ is a weighting parameter.
Note that metrics used in GANs to measure discrepancies between distributions\cite{Arjovsky2017WassersteinGA} may also be considered.
\subsubsection{Regularization Loss $L_{reg}$}
Is defined as
\begin{equation}\label{eq:Lreg}
L_{reg}(x, g_s(x), g_a(x)) =  \lambda 
                \sum_{i \in \mathbb{R}^d} \left \|\bigtriangledown  \left( g^i_s(x) -  g^i_a(x)  \right ) \right  \|_{2}
\end{equation}
where parameter $\lambda \in \mathbb{R}$ controls the relative importance of $L_{reg}$ with respect to the other terms of $L$~\eqref{eq:loss} and $d$ is the dimension of the output space of the generators.
This term acts as $L_{s,a}$ favoring the proximity of $g_s$ and $g_a$ and regularizes the explanation map~\eqref{eq:explanation}.

\subsection{Multi-class situation}\label{subsec:multi-class}
Weak optimization problem~\eqref{eq:loss} can be adapted to the multi-class problem by modifying $L_{f_c}$ to account for a vector valued $\mathbf{f_c}  = [f_{c_i}]_{i \in [|1,\cdots N |]}$ function.
This boils down to modifying term~\eqref{eq:Lfc} adapting CW loss of\cite{Elliott2019AdversarialPO, zhang_generating_2019} into
\begin{equation}\label{eq:Lfc-multiclass}
L_{f_c} = 
    \begin{array}{l}
        \beta_1  \max(\underset{ i\neq l}{\max} (f_{c_i}(g_s(x))) - f_{c_l}(g_s(x)), -\kappa) + \\
        \beta_2 \max(f_{c_l}(g_a(x)) - \underset{ i\neq l}{\max} (f_{c_i}(g_a(x))), - \kappa)
    \end{array}
\end{equation}
where index $l$ is defined by $ \underset{i}{\text{argmax}}([f_{c_i}(x)])$ corresponding to the class selected by the classifier on input $x$. $\kappa$ is a strictly positive margin.

\subsection{Explanation and augmentations}\label{subsec:augmentation}
As our visual explanation is defined as the difference between two generated images, we suggest to regularize the output of our explanation method by averaging all outputs on random geometrical transformations of the input image. Thus, discriminative regions against reconstruction errors are further enforced.
This average reads:
\begin{equation}\label{eq:explanation-augmentation}
    \overline{E}_{f_c}(x) = \frac{1}{N + 1} \left [ E_{f_c}(x) +
    \sum_{i=1}^N \psi_i^{-1} \left({E}_{f_c}(\psi_i(x) ) \right)         \right ]
\end{equation}
where $\psi_i$ are random geometric transformations such as rotations, translations, zoom, axis flip, etc. This particular regularization can be applied to all other visual explanation techniques (see section \ref{subsec:weak-localization-evaluation}).\\

In the following sections, we denote by $x_s = \bar{g}_s(x)$, $x_a = \bar{g}_a(x)$ the output of similar and adversarial generators respectively.
\section{Experiments}\label{sec:experiments}

\subsection{Datasets}\label{subsec:datasets}
We tested our approach on a publicly available Chest X-rays dataset for a binary classification problem.
The Chest X-rays dataset comes from the RSNA Pneumonia Detection Challenge dataset which is a subset of 26,684 exams in dicom taken from the NIH CXR14 dataset\cite{DBLP:journals/corr/WangPLLBS17}.
We only extracted healthy and pneumonia cases from the original dataset.
The resulting database is composed of 14,863 exams: 6,012 pneumonia - 8,851 healthy.
We split the dataset into 3 random groups (80\%, 10\%, 10\%) : train (11,917) - validation (1,495) - test (1,451).
X-rays exams with opacities contain bounding box ground truth annotations.

\subsection{Classifier Set Up}\label{subsec:classifier-set-up}
The classification model whose decisions need to be explained consists of a ResNet50\cite{DBLP:journals/corr/HeZR016}.
We adapt the last layers of the ResNet50 network in order to tackle a binary classification task (healthy/pathology).
We transfer the pre-trained backbone layers from Imagenet\cite{imagenet_cvpr09} to our binary classifier.
Then, the network is trained on the whole training set for 50 epochs with a batch size of 32.
We use the Adam optimizer\cite{DBLP:journals/corr/KingmaB14} with an initial learning rate of 1e-4.
Original X-rays are resized from 1024x1024 to 224x224 and normalized to $[0,1]$.
We also used zoom, translations, rotations and vertical flips as random data augmentations. The binary classifier achieves an AUC of 0.974 on the test set.

\subsection{Generative Explanation Model}\label{subsec:generator_imple}
For the similar and adversarial generators, as in\cite{pix2pix2017, zhang_generating_2019}, generators roughly follows the UNet architecture\cite{DBLP:journals/corr/RonnebergerFB15}.
We propose two different types of generators:
(i) \textit{Duo $AE$} (Figure~\ref{fig:duo_single_gen} - Top): $g_s$ and $g_a$ are two separated UNet auto-encoders. (ii) \textit{Single $AE_i$} (Figure~\ref{fig:duo_single_gen} - Bottom): $g_s$ and $g_a$ share a same auto-encoder part that captures image structure for both generators.
They differ by two identical convolutional neural networks connected at the end of the common autoencoder. Index $i$ indicates the number of convolutional layers in the separated CNN.

Generators take as input the same image as the classifier with 3 channels and dimensions 224x224.
Both generators are trained simultaneously for 70 epochs with a batch size of 8 for \textit{Single $AE_i$} and 4 for \textit{Duo $AE$ }, with the same augmentations used for the classifier.
Adam optimizer is used with an initial learning rate of 1e-4, and we reduce the learning rate by 3 each time the loss does not decrease after 3 epochs. 
Through trial an error we selected the objective loss function \eqref{eq:loss} parameters providing the best results and summarized them in Table~\ref{tab:params}. 

\begin{table}[htbp]
\caption {Selected parameters for couples of generators}
\begin{center}
\begin{tabular}{cccccccc}
    \hline
    \textbf{Model name} & $\alpha_1$ & $\alpha_2$ &   $\alpha_3$ & $\alpha_4$ & $\beta_{1,2}$ & $\gamma$ & $\lambda$ \\
    \hline\hline
    Duo $AE$ (TV)&1&1&1&0&0.001&0&0.2\\
    Duo $AE$ (W,TV)&1&1&1&0&0.001&0.1&0.2\\
    Single $AE_1$ (TV)&1&1&1&0&0.001&0&0.2\\
    Single $AE_1$ (W)&3&1&1&0.2&0.001&0.1&0\\
    Single $AE_1$ (W,TV)&1&1&1&0.2&0.001&0.1&0.2\\
    Single $AE_2$ (W)&3&1&1&0.2&0.001&0.2&0\\
    Single $AE_2$ (W, TV)&3&1&1&0.2&0.001&0.2&0.2\\
    \hline
\end{tabular}
\label{tab:params}
\end{center}
\end{table}

\subsection{Augmentation during generator's prediction}\label{subsec:augmentation-during-generator's-prediction}
During generator's prediction, for each image $x$, we generate 10 augmented images $(x_i)_{i \in \left [ \left | 1, 10 \right |  \right ] }$ with random geometric transformations of parameters described in Table~\ref{tab:aug_params}

\begin{table}[htbp]
\caption {Augmentation Parameters}
\begin{center}
\begin{tabular}{cc}
    \hline
    \textbf{Transformation} & value(s)  \\
    \hline\hline
    Rotations range ($^{\circ})$& $[-5, 5]$\\
    Height shift range (pixels) &$[-10, 10]$\\
    Width shift range (pixels) &$[-10, 10]$\\
    Zoom range & $[0.9, 1]$\\
    Random horizontal flip & (True, False)\\
    Random vertical flip&(True, False)\\
    \hline
\end{tabular}
\label{tab:aug_params}
\end{center}
\end{table}


\subsection{Method Evaluation}\label{subsec:evaluations}

\noindent\textbf{Generators Evaluation}\\
The evaluation is achieved on the classifier's test set.
For similar and adversarial generators $\bar{g}_s$ and $\bar{g}_a$, we respectively evaluate the similarity between $x$, $x_s$ and $x_a$.
Structural Similarity Index (SSIM), as well as the Peak Signal to Noise Ratio (PSNR) are used to evaluate the similarity between pair of images. For the classification purpose, we compute the area under the ROC curve between the rounded value of the classifier predictions $R(f_c(x))$ (resp. $R(1-f_c(x))$) and $f_c(x_s)$ (resp. $f_c(x_a)$).
\\

\noindent\textbf{Interpretability Evaluation}\\
In state of the art methods and more specifically in medical imaging, a visual explanation is considered as interpretable if:
(i) The highlighted regions coincide with discriminative regions for humans.
In our classification problem, salient regions should overlap opacity regions where the pathologies are found.
(ii) The highlighted regions coincide with context regions that are also discriminative for humans.
We can quantitatively assess the overlap between explanation map and ground truth annotations by conducting a weak localization experiment.
We use two metrics to evaluate the localization performance: the intersection over union ($IOU$) and an estimated area under the curve ($AUC_{Loc}$).
We compute the intersection over union between the ground truth mask $M_{GT}$ and the thresholded explanation mask $M_{Ei}$, as defined in~\eqref{eq:iou_i}.
\begin{equation}\label{eq:iou_i}
    IoU_i = \frac{M_{GT}\cap M_{Ei}}{M_{GT}\cup M_{Ei}}
\end{equation}
where $M_{GT}$ is the binary mask included inside the ground truth bounding box annotation, and $M_{Ei}$ is the binary mask obtained when we threshold the explanation mask $E_{f_c}$ at the $i$-th percentile $p_i$:
\begin{equation}\label{eq:mei}
    M_{Ei} = \left\{\begin{matrix}
1 & E_{f_c} \geq p_i \\
0 & \text{otherwise}
\end{matrix}\right.
\end{equation}
We also measure the precision and the sensitivity of the localization for different thresholds $p_i$ in order to compute the area under the precision and recall curve as introduced in\cite{fu2020distributionguided}:
\begin{equation}\label{eq:auc}
    AUC_{Loc} = \sum_i P_i (R_i - R_{i-1})
\end{equation}
where $P_i = \frac{M_{GT}\cap M_{Ei}}{M_{Ei}} $, $R_i = \frac{M_{GT}\cap M_{Ei}}{M_{GT}}$ and $i \in \left [ \left | 1, 100 \right |  \right ] $.
Our estimation of $AUC_{Loc}$ differs from\cite{fu2020distributionguided} as we only compute the metrics over the hundred values of percentile instead of all sorted values of the explanation map.\\
We also compute a partial $AUC_{Loc}$ for percentiles between $80$ and $100$ as it is more representative of the volume occupied by the ground truth mask $M_{GT}$.
We show some statistics of the bounding box annotations in Table~\ref{bbox-statistics}.\\
We compare our proposed method to the \textbf{\textit{naive}} one (see section~\ref{subsec:method}) and to the following state of the art approaches: Gradient\cite{simonyan_deep_2014}, Smooth-Gradient\cite{smilkov_smoothgrad:_2017}, Input Gradient\cite{hechtlinger_interpretation_2016}, Integrated Gradient\cite{sundararajan_axiomatic_2017}, GradCAM\cite{selvaraju_grad-cam:_2017}, BBMP\cite{fong_interpretable_2017}, Mask Generator\cite{dabkowski_real_2017} and Perceptual Perturbation \cite{Elliott2019AdversarialPO}.
The best BBMP results are reached when looking for a mask at 56x56 and with Gaussian blur perturbation.
The mask Generator follows the UNet architecture described in\cite{dabkowski_real_2017}, but we remove the class selector and adapt the objective function to a single class problem.
The best results are obtained when we generate a mask at size 112x112 and then upsample it to image dimensions. For Perceptual perturbation which is not model-agnostic, we regularize the first ReLU layer of each convolution block of the ResNet50 classifier. We also adapt the optimization to a single class problem.
\begin{table}[htbp]
\caption {Opacities bounding box statistics}
\begin{center}
\begin{tabular}{cccc}
    \hline
    \textbf{Metrics} & Height (pixels) & Width (pixels) & Area Ratio (\%)  \\
    \hline\hline
    Min & 13 & 13& 0.5\\
    Max& 171&91&25.3\\
    Mean &71.8&47.5&7.3\\
    Median &67.8&46.8&6.3 \\
    \hline
\end{tabular}
\label{bbox-statistics}
\end{center}
\end{table}

\section{Results \& Discussion}\label{sec:results_and_discussion}

\subsection{Generator evaluation}\label{subsec:generator-evaluation}
For the different architectures and optimization tested, both generators $g_s$ and $g_a$ reach high performance in term of classification.
As shown in Table~\ref{tab:classif}, similar images are almost all classified as the original ones, as the $AUC_s$ almost reaches $1$.
Adversarial images achieve better adversarial attacks either when the network (Single) or the weights regularization (W) causes the generators $g_s$ and $g_a$ to be close to each other (Table~\ref{tab:classif}).
They even outperform the \textbf{\textit{naive}} approach (Adv. $AE$ (TV)) where $g_a$ is trained without $g_s$.\\

\begin{figure}[htbp]
\centering
    \includegraphics[width=1\linewidth]{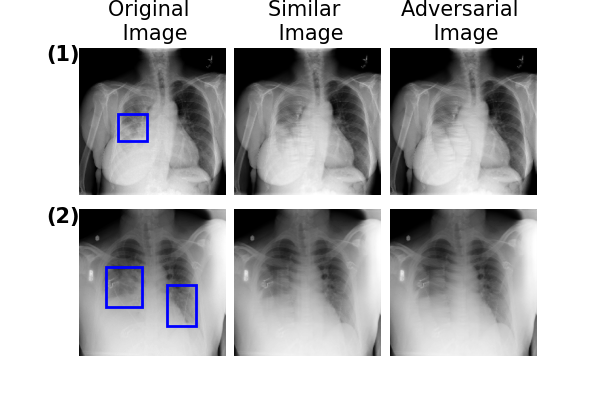}
  \caption{Examples of original images with respective similar and adversarial generated images. \textbf{Case (1)}:$f_c(x)=1.0$ - $f_c(x_s) = 0.999$ - $f_c(x_a)=0.051$ - PSNR between orignal and similar image $PSNR_{os}=43.18$ - PSNR between orignal and adversarial image $PSNR_{oa}=42.07$ - PSNR between similar and adversarial image $PSNR_{sa}=52.75$. \textbf{Case (2)}: $f_c(x)=0.978$ - $f_c(x_s) = 0.986$ - $f_c(x_a)=0.34$ -  $PSNR_{os}=46.30$ - $PSNR_{oa}=45.49$ - $PSNR_{sa}=56.39$}
  \label{fig:adv_sim_reconstructions}
\end{figure}

\begin{table}[htbp]
\caption {Classification AUC on similar and adversarial images}
\begin{center}
\begin{tabular}{ccc}
    \hline
    \textbf{Explanation method}&$AUC_s$&$AUC_a$\\
    \hline\hline
    Naive & - & 0.939\\
    Duo $AE$ (TV)& 1.0&0.905\\
    Duo $AE$ (W,TV)&1.0 &0.958\\
    Single $AE_1$ (TV)& 1.0 & 0.961\\
    Single $AE_1$ (W) & 0.998&0.952\\
    Single $AE_1$ (W,TV)& 0.997&0.944\\
    Single $AE_2$ (W)& 0.998&0.949\\
    Single $AE_2$ (W, TV)&0.998 &0.952\\
    \hline
\end{tabular}
\label{tab:classif}
\end{center}
\end{table}
For the similarity, both generators produce samples visually highly similar to original images (see Figure~\ref{fig:adv_sim_reconstructions} and Table~\ref{tab:similarity}).
Similar images $x_s$ best perform for both SSIM and PSNR when generators are not constrained by weight regularization.
At the opposite, adversarial images $x_a$ increase their similarity to both $x$ and $x_s$ when generators are constrained, and it even outperforms the \textbf{\textit{naive}} adversarial generator trained on its own.
In our case, the objective is to produce $x_s$ and $x_a$ as close as possible in order to reduce non discriminative differences, while having $x_s$ very close to $x$.
As shown in Table~\ref{tab:similarity}, Single $AE_2$ regularized with weights proximity (W) produce highly similar samples $x_s$ and $x_a$, while maintaining a strong similarity between $x_s$ and $x$.

\begin{table}[htbp]
\caption {Similarity metrics between generated and original images}
\begin{center}
\begin{tabular}{ccccccc}
    \hline
    \textbf{Explanation method}&\multicolumn{2}{c}{$x \leftrightarrow x_s$}& \multicolumn{2}{c}{$x \leftrightarrow x_a$} & \multicolumn{2}{c}{$x_s \leftrightarrow x_a$}\\
    \textit{Metrics}&ssim&psnr&ssim&psnr&ssim&psnr\\
    \hline\hline
    Adv. $AE$ (TV) & - &-&0.994&41.92&-&-\\
    Duo $AE$ (TV) &0.996&44.07&0.987&39.47&0.994&43.89\\
    Duo $AE$ (W,TV) &0.995&41.99&0.987&39.08&0.995&44.26\\
    Single $AE_1$ (TV)& \textbf{0.997}&\textbf{44.57}&0.989&40.67&0.996&45.25\\
    Single $AE_1$ (W) & 0.994&42.73&0.993&41.85&0.999&52.59\\
    Single $AE_1$ (W,TV)& 0.992&41.79&0.991&41.35&\textbf{0.999}&\textbf{54.55}\\
    Single $AE_2$ (W) &0.995&43.61&0.994&42.42&0.999&52.26\\
    Single $AE_2$ (W, TV) &0.995&43.88&\textbf{0.994}&\textbf{42.63}&0.999&51.93\\
    \hline
\end{tabular}
\label{tab:similarity}
\end{center}
\end{table}

\subsection{Weak localization evaluation}\label{subsec:weak-localization-evaluation}
As shown in Table~\ref{bbox-statistics}, bounding box annotations of the test set occupy from 0.5 to 25.3 \% of the image with an average occupation of 7.3\%.
For different generators and regularizations, we accordingly list the results of the averaged IOU for $p_i$ between the 80th and 100th percentile value in Table~\ref{tab:iou_gen}, and total and partial $AUC_{Loc}$ in Table~\ref{tab:auc_gen}.
Firstly, Single $AE$ clearly outperforms the Duo version for all IOU and AUC scores.
The Single $AE$ approach compelled $g_a$ and $g_s$ to capture the same information on the original image by sharing a common autoencoder.
As shown in Table~\ref{tab:similarity}, the proximity between $x$ and $x_a$ as well as between $x_s$ and $x_a$ is better for Single $AE$ approaches.
\begin{table}[htbp]
\caption {IOU scores at different thresholds of binarization - Comparison across the different generators architectures}
\begin{center}
\begin{tabular}{cccccc}
    \hline
    \textbf{Explanation method}&\multicolumn{5}{c}{\textbf{IOU}}\\
    \textit{Percentile}&80&85&90&95&98\\
    \hline\hline
    Duo $AE$ (TV)& 0.190&0.182&0.164&0.122&0.070\\
    Duo $AE$ (W,TV)& 0.188&0.184&0.170&0.132&0.079\\
    Single $AE_1$ (TV)& 0.187&0.182 &0.166&0.127&0.075\\
    Single $AE_1$ (W) & 0.227&0.222&0.204&0.157&0.090\\
    Single $AE_1$ (W,TV)& 0.234&0.235&0.220&0.171&\textbf{\textit{0.099}}\\
    Single $AE_2$ (W)& 0.240&0.245&0.229&0.172&0.095\\
    Single $AE_2$ (W, TV)& \textbf{\textit{0.248}}&\textbf{\textit{0.250}}&\textbf{\textit{0.232}}&\textbf{\textit{0.173}}&0.097\\
    \hline
    \textbf{\textit{With Augmentations}}&&&\\
    Duo $AE$ (TV)& 0.243&0.232 & 0.206&0.156&0.085\\
    Duo $AE$ (W,TV)&0.263&0.253& 0.227&0.166&0.093\\
    Single $AE_1$ (TV)& 0.262&0.249 &0.218&0.156&0.086\\
    Single $AE_1$ (W) & 0.262&0.254&0.233&0.181&0.105\\
    Single $AE_1$ (W,TV)&0.268&0.261& 0.240&0.188&0.112\\
    Single $AE_2$ (W)&0.288 &0.288&0.268&0.204&\textbf{0.115}\\
    Single $AE_2$ (W, TV)& \textbf{0.292} &\textbf{0.292}&\textbf{0.272}&\textbf{0.206}&\textbf{0.115}\\
    \hline
\end{tabular}
\label{tab:iou_gen}
\end{center}
\end{table}
 Then, the weights regularization between similar path and adversarial path introduced in~\eqref{eq:gw} improves all the localization performance e.g. from $IOU_{90} = 0.166$ to $IOU_{90} = 0.220$ for Single $AE_1$ (TV).
This is consistent with the findings in~\ref{tab:similarity}.
Total variation regularization on the resulting explanation mask also slightly increases IOU and AUC scores for Single $AE_{1,2}$.
In addition, the Single generator with two convolutional layers ($AE_2$) performs better than the single-layer one ($AE_1$).\\
Finally, the use of augmentations during generator's prediction improves localization scores for all cases e.g. up to 4 points for $IOU_{90}$ (Table~\ref{tab:iou_gen}), from 7 to 11 points for total and partial $AUC_{Loc}$ (Table~\ref{tab:auc_gen}).
\begin{table}[htbp]
\caption {Estimated AUC scores for Precision-Recall - Comparison across the different generators architectures}
\begin{center}
\begin{tabular}{cccc}
    \hline
    \textbf{Explanation method}& \textbf{Total AUC} & \textbf{Partial AUC}\\
    \hline\hline
    Duo $AE$ (W,TV)&0.257&0.162\\
    Single $AE_1$ (TV)& 0.253&0.157\\
    Single $AE_1$ (W)& 0.310&0.220\\
    Single $AE_1$ (W, TV)& 0.325&0.239\\
    Single $AE_2$ (W)& 0.325&0.248\\
    Single $AE_2$ (W, TV)& \textbf{0.339} &\textbf{0.256}\\
    \hline
    \textbf{\textit{With Augmentations}}&&&\\
    Duo $AE$ (W,TV)& 0.362& 0.263 \\
    Single $AE_1$ (TV)& 0.353 & 0.254\\
    Single $AE_1$ (W)& 0.370&0.274\\
    Single $AE_1$ (W,TV)& 0.381&0.287\\
    Single $AE_2$ (W)& 0.405&0.322\\
    Single $AE_2$ (W, TV)& \textbf{0.412} &\textbf{0.328}\\
    \hline
\end{tabular}
\label{tab:auc_gen}
\end{center}
\end{table}

\begin{figure}[htbp]
\centering
\makebox[\linewidth][c]{\includegraphics[width=1.05\linewidth]{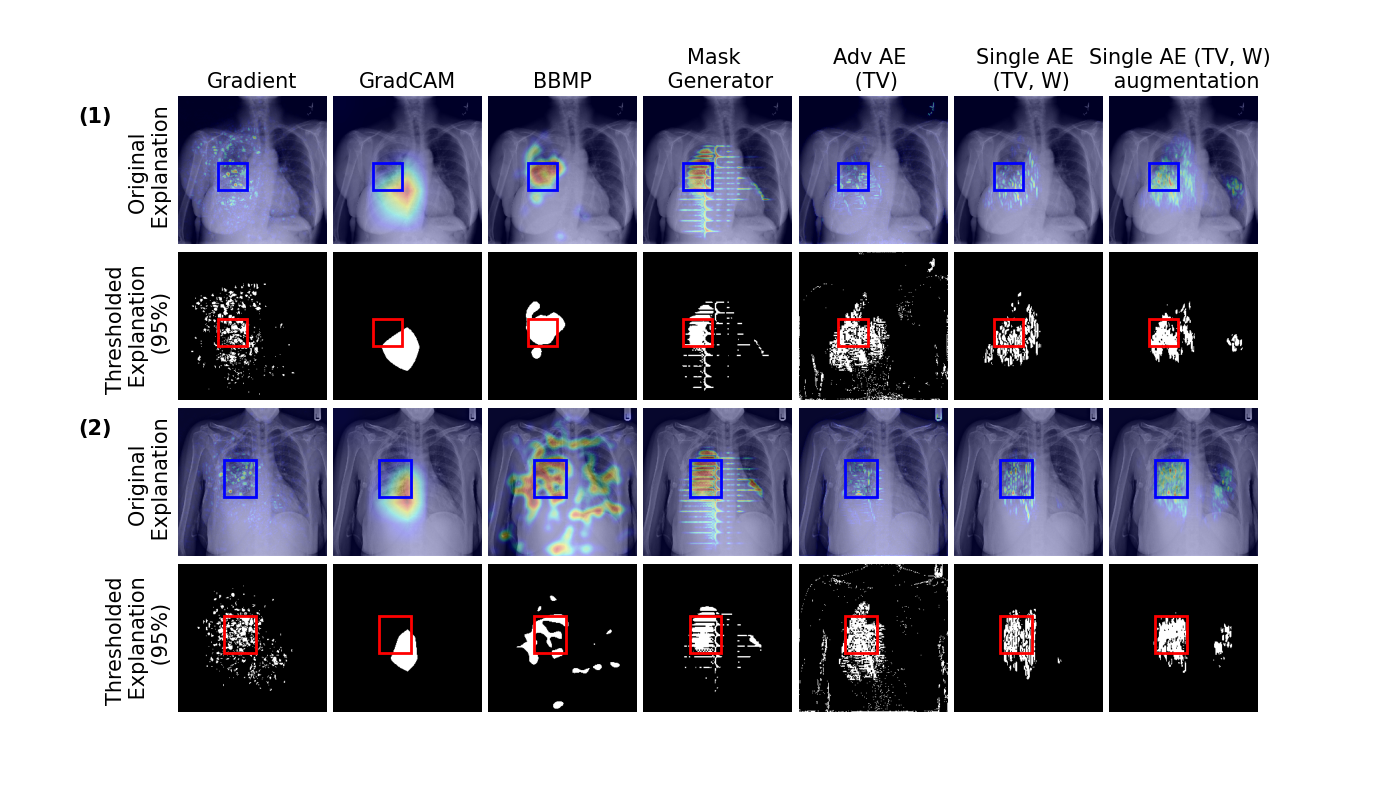}}%
  \caption{Examples of explanation maps generated by different methods in case of a single ground truth bounding box annotation. \textit{Top row}: the original image with the explanation map and the ground truth bounding box. \textit{Bottom row}: Binary heatmaps for the $95^{th}$ percentile}
  \label{fig:single_box_cases}
\end{figure}

\begin{figure}[htbp]
\centering
    \makebox[\linewidth][c]{\includegraphics[width=1.05\linewidth]{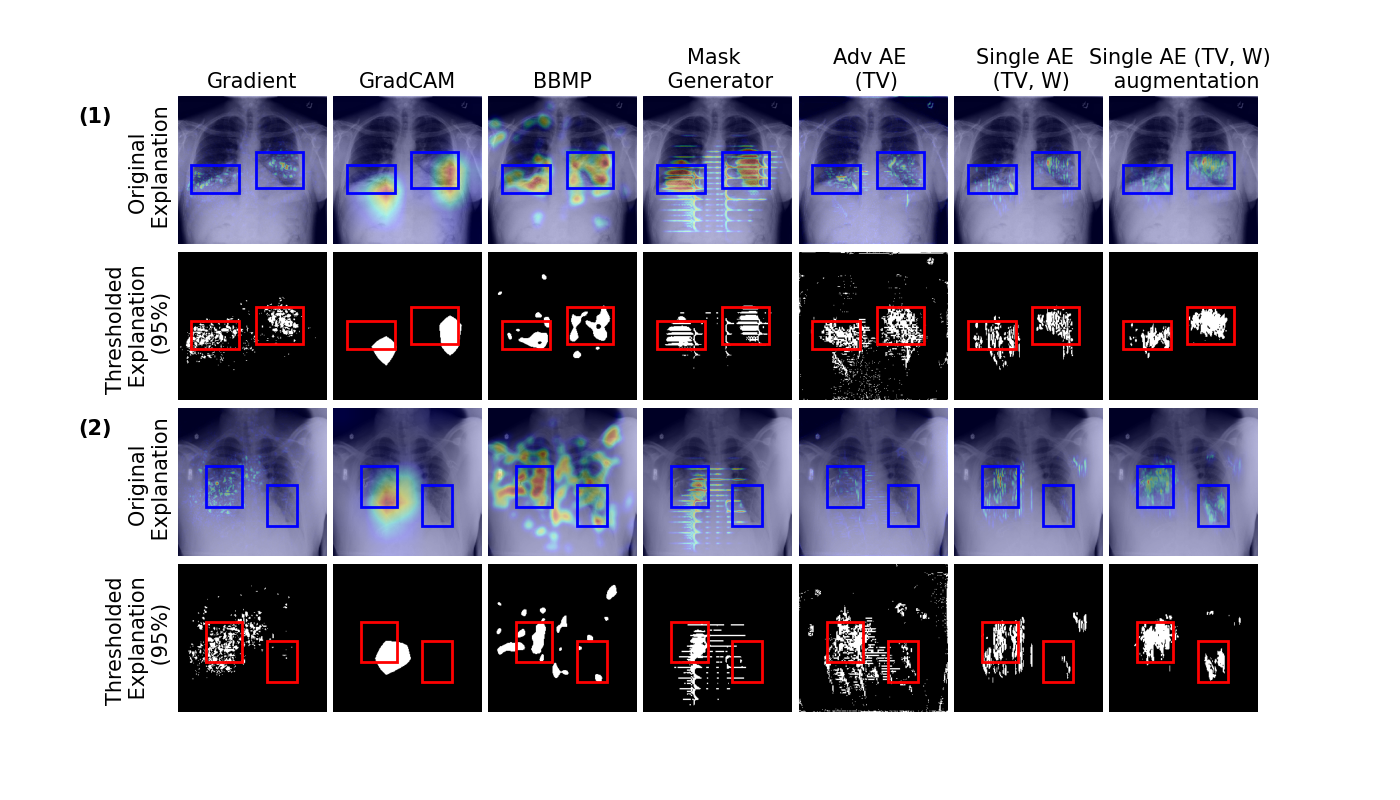}}
  \caption{Explanation maps generated by different methods in case of two ground truth bounding box annotations. \textit{Top row}: the original image with the explanation map and the ground truth bounding box. \textit{Bottom row}: Binary heatmaps for the $95^{th}$ percentile}
  \label{fig:double_box_cases}
\end{figure}

When compared to state of the art methods (Tables~\ref{tab:iou_soa},~\ref{tab:auc_soa}), Single $AE_2$ (W, TV) achieves comparable localization scores.
Our method even slightly outperforms the best performers Mask Generator and BBMP for IOU scores for percentile thresholds from 80 to 95 \%.
It is also the case for both partial and total AUC compared to the best state of the art approaches: GradCAM, BBMP and Mask Generator.
Only Mask Generator and Gradient outperform or compete with our method for $IOU_{98}$.
We can also note that the \textbf{\textit{naive}} explanation directly defined as the difference between $x_a$ and $x$ (Adv. $AE$ (TV)) produces much poorer results.\\
However, when using augmentation during generator prediction phase, our method outperforms all the others. Visual illustrations are given in Figures~\ref{fig:single_box_cases} and~\ref{fig:double_box_cases} for cases where the opacities are located either at one or two different positions. When thresholding heatmaps at the 95th percentile, our method (\textit{Single $AE$}) seems to generate less noisy masks than other approaches including the \textbf{\textit{naive}} one (\textit{Adv $AE$}), while capturing all discriminative structures.
In addition, our method is suitable for real time situation as suggests the generation time per image of the explanation given in Table~\ref{tab:auc_soa} (on NVIDIA GPU MX130).
\begin{table}[htbp]
\caption {IOU scores at different thresholds of binarization - Comparison to State of the Art Methods}
\begin{center}
\begin{tabular}{cccccc}
    \hline
    \textbf{Explanation method}&\multicolumn{5}{c}{\textbf{IOU}}\\
    \textit{Percentile}&80&85&90&95&98\\
    \hline\hline
    Gradient&0.203&0.199&0.187&0.152&0.097\\
    Smooth Grad. & 0.192&0.188&0.176&0.143&0.091\\
    Input Grad. & 0.191&0.185&0.170&0.136&0.086\\
    Integrated Grad.& 0.176&0.171&0.157&0.124&0.077\\
    GradCAM&0.237&0.225&0.195&0.138&0.070\\
    BBMP&0.233&0.226&0.204&0.154&0.087\\
    \hline
    Perceptual Perturbation&0.133&0.125&0.110&0.080&0.045\\
    \hline
    Mask Generator& 0.222&0.219&0.208&0.169&\textbf{\textit{0.103}}\\
    \hline
    Adv. $AE$ (TV) & 0.177&0.173&0.158&0.118&0.064\\
    \hline
    \textbf{\textit{Adversarial vs Similar}}&&&\\
    Single $AE_2$ (W, TV)& \textbf{\textit{0.248}}&\textbf{\textit{0.250}}&\textbf{\textit{0.232}}&\textbf{\textit{0.173}}&0.097\\
    \hline
    \textbf{\textit{Adv. vs Sim. + Augment.}}&&&\\
    Single $AE_2$ (W, TV)& \textbf{0.292} &\textbf{0.292}&\textbf{0.272}&\textbf{0.206}&\textbf{0.115}\\
    \hline
\end{tabular}
\label{tab:iou_soa}
\end{center}
\end{table}

\begin{table}[htbp]
\caption {Estimated AUC scores for Precision-Recall and Computation time - Comparison to State of the Art Methods}
\begin{center}
\begin{tabular}{ccccc}
    \hline
    \textbf{Explanation method}& \textbf{Total AUC} & \textbf{Partial AUC} &\textbf{Time (s)}\\
    \hline\hline
    Gradient&0.287&0.189&2.04\\
    Integrated Grad.& 0.244&0.146&1.93\\
    GradCAM&0.324&0.235&0.78\\
    BBMP&0.326&0.229&17.14\\
    \hline
    Perceptual Perturbation&0.180&0.084&30.74\\
    \hline
    Mask Generator& 0.327&0.226&0.09\\
    \hline
    Adv. $AE$ (TV) & 0.238&0.145&0.10\\
    \hline
    \textbf{\textit{Adversarial vs Similar}}& &\\
    Single $AE_2$ (W, TV)& \textbf{\textit{0.339}} &\textbf{\textit{0.256}}&0.05\\
    \hline
    \textbf{\textit{Adv. vs Sim. + Augment.}}&&&\\
    Single $AE_2$ (W, TV)& \textbf{0.412} &\textbf{0.328}&0.63\\
    \hline
\end{tabular}
\label{tab:auc_soa}
\end{center}
\end{table}
As an additional experiment, we apply the augmentation technique to other state of the art methods that produce their visual explanation in one shot.
Localization results are listed in Tables~\ref{tab:iou_soa_aug} and~\ref{tab:auc_soa_aug}.
All localization scores improve, while the generation time per image remains adequate (see Table ~\ref{tab:auc_soa_aug}).
By using augmentations, we observe for all methods a gain similar to that observed for our method.
Our best method still achieves better localization results for AUC metrics.
For IOU, Mask Generator outperforms our method for  $p_i \geq p_{95}$.
\begin{table}[htbp]
\caption {IOU scores at different thresholds of binarization -  Comparison to State of the Art Methods without (\textbf{Top}) and with (\textbf{Bottom}) augmentations}
\begin{center}
\begin{tabular}{cccccc}
    \hline
    \textbf{Explanation method}&\multicolumn{5}{c}{\textbf{IOU}}\\
    \textit{Percentile}&80&85&90&95&98\\
    \hline\hline
    \multirow{2}{*}{Gradient [1]}&0.203&0.199&0.187&0.152&0.097\\
    &\textbf{\textit{0.256}}&\textbf{\textit{0.252}}&\textbf{\textit{0.236}}&\textbf{\textit{0.190}}&\textbf{\textit{0.117}}\\
    \hline
    \multirow{2}{*}{GradCAM [2]}&0.237&0.225&0.195&0.138&0.070\\
    &\textbf{\textit{0.271}}&\textbf{\textit{0.263}}&\textbf{\textit{0.244}}&\textbf{\textit{0.190}}&\textbf{\textit{0.105}}\\
    \hline
    BBMP [3]&0.233&0.226&0.204&0.154&0.087\\
    \hline
    \multirow{2}{*}{Mask Generator [4]}& 0.222&0.219&0.208&0.169&0.103\\
    &\textbf{\textit{0.259}}&\textbf{\textit{0.264}}&\textbf{\textit{0.259}}&\textbf{\textcolor{red}{0.221}}&\textbf{\textcolor{red}{0.137}}\\
    \hline
    \multirow{2}{*}{"Naive"}  &0.177&0.173&0.158&0.118&0.064\\
     &\textbf{\textit{0.239}}&\textbf{\textit{0.230}}&\textbf{\textit{0.208}}&\textbf{\textit{0.156}}&\textbf{\textit{0.087}}\\
    \hline
    \multirow{2}{*}{\textbf{Ours}}&0.248 &0.250&0.232&0.173&0.097\\
    &\textbf{\textcolor{red}{0.292}} &\textbf{\textcolor{red}{0.292}}&\textbf{\textcolor{red}{0.272}}&\textbf{\textit{0.206}}&\textbf{\textit{0.115}}\\
    \hline
\end{tabular}
\label{tab:iou_soa_aug}
\end{center}
\end{table}

\begin{table}[htbp]
\caption {Estimated AUC scores for Precision-Recall and Computation time -  Comparison to State of the Art Methods without (\textbf{Top}) and with (\textbf{Bottom}) augmentations}
\begin{center}
\begin{tabular}{ccccc}
    \hline
    \textbf{Explanation method}& \textbf{Total AUC} & \textbf{Partial AUC} &\textbf{Time (s)}\\
    \hline\hline
    \multirow{2}{*}{Gradient [1]}&0.287&0.189&2.04\\
    &\textbf{\textit{0.374}}&\textbf{\textit{0.274}}&2.83\\
    \hline
    \multirow{2}{*}{GradCAM [2]}&0.326&0.235&0.78\\
    &\textbf{\textit{0.397}}&\textbf{\textit{0.302}}&5.09\\
    \hline
    BBMP [3]&0.326&0.229&17.14\\
    \hline
    \multirow{2}{*}{Mask Generator [4]} &0.327&0.226&0.09\\
    &\textbf{\textit{0.404}}&\textbf{\textit{0.308}}&0.68\\
    \hline
    \multirow{2}{*}{"Naive"} & 0.238&0.145&0.10\\
    & \textbf{\textit{0.325}}&\textbf{\textit{0.232}}&0.75\\
    \hline
    \multirow{2}{*}{\textbf{Ours}}& 0.339 &0.256&0.05\\
    &\textbf{\textcolor{red}{0.412}} &\textbf{\textcolor{red}{0.328}}&0.63\\
    \hline
\end{tabular}
\label{tab:auc_soa_aug}
\end{center}
\end{table}
\section{Conclusion}\label{sec:conclusion}
In this work, we introduce a new method to produce a visual explanation of the classifier's decision that leverages adversarial generation learning.
We propose to train simultaneously a couple of generators to produce an adversarial image that goes against the classifier's decision, and a similar image that is classified as the original one.
We show that the differences between the two images as well as the learning procedure helps to better capture discriminative features. We have tested our method on a binary classification problem in the medical domain.
We have shown that our method outperforms
state of the art techniques in terms of weak localization, especially when we introduced geometric augmentations during the generation phase. Unlike some state of the art methods, our proposed method is both model-agnostic and sufficient for real time situation such as medical image analysis. Finally, we show that
random geometric augmentations applied to the original image improves all the tested state of the art approaches.\\
In future works, we shall generalize our method to multi-classification problems and apply it to 3D medical image problems.

\bibliographystyle{IEEEtran}\
\bibliography{state_of_the_art,sanity,perturbation_based,adversarial_generation,adversarial_training,models,datasets}

\begin{thebibliography}{10}
\providecommand{\url}[1]{#1}
\csname url@samestyle\endcsname
\providecommand{\newblock}{\relax}
\providecommand{\bibinfo}[2]{#2}
\providecommand{\BIBentrySTDinterwordspacing}{\spaceskip=0pt\relax}
\providecommand{\BIBentryALTinterwordstretchfactor}{4}
\providecommand{\BIBentryALTinterwordspacing}{\spaceskip=\fontdimen2\font plus
\BIBentryALTinterwordstretchfactor\fontdimen3\font minus
  \fontdimen4\font\relax}
\providecommand{\BIBforeignlanguage}[2]{{%
\expandafter\ifx\csname l@#1\endcsname\relax
\typeout{** WARNING: IEEEtran.bst: No hyphenation pattern has been}%
\typeout{** loaded for the language `#1'. Using the pattern for}%
\typeout{** the default language instead.}%
\else
\language=\csname l@#1\endcsname
\fi
#2}}
\providecommand{\BIBdecl}{\relax}
\BIBdecl

\bibitem{skincancerEsteva2017}
A.~Esteva, B.~Kuprel, R.~Novoa, J.~Ko, S.~Swetter, H.~Blau, and S.~Thrun,
  ``Dermatologist-level classification of skin cancer with deep neural
  networks,'' in \emph{Nature}, vol. 542, 2017, pp. 115–--118.

\bibitem{Holzinger2017WhatDW}
A.~Holzinger, C.~Biemann, C.~S. Pattichis, and D.~B. Kell, ``What do we need to
  build explainable ai systems for the medical domain?'' \emph{ArXiv}, vol.
  abs/1712.09923, 2017.

\bibitem{simonyan_deep_2014}
K.~Simonyan, A.~Vedaldi, and A.~Zisserman, ``Deep {Inside} {Convolutional}
  {Networks}: {Visualising} {Image} {Classification} {Models} and {Saliency}
  {Maps},'' in \emph{ICLR}, 2014.

\bibitem{smilkov_smoothgrad:_2017}
D.~Smilkov, N.~Thorat, B.~Kim, F.~B. Vi{\'e}gas, and M.~Wattenberg,
  ``Smoothgrad: removing noise by adding noise,'' \emph{ArXiv}, vol.
  abs/1706.03825, 2017.

\bibitem{hechtlinger_interpretation_2016}
Y.~Hechtlinger, ``Interpretation of prediction models using the input
  gradient,'' \emph{ArXiv}, vol. abs/1611.07634, 2016.

\bibitem{sundararajan_axiomatic_2017}
M.~Sundararajan, A.~Taly, and Q.~Yan, ``Axiomatic attribution for deep
  networks,'' in \emph{ICML}, 2017.

\bibitem{zhou_learning_2016}
B.~Zhou, A.~Khosla, {\`A}.~Lapedriza, A.~Oliva, and A.~Torralba, ``Learning
  deep features for discriminative localization,'' in \emph{CVPR}, 2016, pp.
  2921--2929.

\bibitem{selvaraju_grad-cam:_2017}
R.~R. Selvaraju, M.~Cogswell, A.~Das, R.~Vedantam, D.~Parikh, and D.~Batra,
  ``Grad-{CAM}: {Visual} {Explanations} from {Deep} {Networks} via
  {Gradient}-{Based} {Localization},'' in \emph{{ICCV}}, 2017, pp. 618--626.

\bibitem{fong_interpretable_2017}
R.~C. {Fong} and A.~{Vedaldi}, ``Interpretable explanations of black boxes by
  meaningful perturbation,'' in \emph{ICCV}, 2017, pp. 3449--3457.

\bibitem{dabkowski_real_2017}
P.~Dabkowski and Y.~Gal, ``Real time image saliency for black box
  classifiers,'' in \emph{NIPS}, 2017, pp. 6967--6976.

\bibitem{goodfellow_explaining_2014}
I.~J. Goodfellow, J.~Shlens, and C.~Szegedy, ``Explaining and {Harnessing}
  {Adversarial} {Examples},'' in \emph{ICLR}, 2015.

\bibitem{Elliott2019AdversarialPO}
A.~Elliott, S.~Law, and C.~Russell, ``Adversarial perturbations on the
  perceptual ball,'' \emph{ArXiv}, vol. abs/1912.09405, 2019.

\bibitem{xiao_generating_2019}
C.~Xiao, B.~Li, J.-Y. Zhu, W.~He, M.~Liu, and D.~X. Song, ``Generating
  adversarial examples with adversarial networks,'' in \emph{IJCAI}, 2018.

\bibitem{zhang_generating_2019}
W.~Zhang, ``Generating adversarial examples in one shot with image-to-image
  translation gan,'' in \emph{IEEE Access}, vol.~7, 2019, pp.
  151\,103--151\,119.

\bibitem{zeiler_visualizing_2013}
M.~D. Zeiler and R.~Fergus, ``Visualizing and understanding convolutional
  networks,'' in \emph{ECCV}, 2014.

\bibitem{springenberg_striving_2014}
J.~T. Springenberg, A.~Dosovitskiy, T.~Brox, and M.~A. Riedmiller, ``Striving
  for simplicity: The all convolutional net,'' in \emph{ICLR}, vol.
  abs/1412.6806, 2015.

\bibitem{rajpurkar_chexnet:_2017}
P.~Rajpurkar, J.~Irvin, K.~Zhu, B.~Yang, H.~Mehta, T.~Duan, D.~Y. Ding,
  A.~Bagul, C.~P. Langlotz, K.~S. Shpanskaya, M.~P. Lungren, and A.~Y. Ng,
  ``Chexnet: Radiologist-level pneumonia detection on chest x-rays with deep
  learning,'' \emph{ArXiv}, vol. abs/1711.05225, 2017.

\bibitem{adebayo_sanity_2018}
\BIBentryALTinterwordspacing
J.~Adebayo, J.~Gilmer, M.~Muelly, I.~Goodfellow, M.~Hardt, and B.~Kim,
  ``\BIBforeignlanguage{en}{Sanity {Checks} for {Saliency} {Maps}},''
  \emph{\BIBforeignlanguage{en}{arXiv:1810.03292 [cs, stat]}}, Oct. 2018,
  arXiv: 1810.03292. [Online]. Available: \url{http://arxiv.org/abs/1810.03292}
\BIBentrySTDinterwordspacing

\bibitem{Rebuffi2020ThereAB}
S.-A. Rebuffi, R.~Fong, X.~Ji, and A.~Vedaldi, ``There and back again:
  Revisiting backpropagation saliency methods,'' in \emph{CVPR}, 2020.

\bibitem{ribeiro_why_2016}
M.~T. Ribeiro, S.~Singh, and C.~Guestrin, ``"why should i trust you?":
  Explaining the predictions of any classifier,'' in \emph{ACM SIGKDD}, 2016.

\bibitem{chang_explaining_2019}
C.-H. Chang, E.~Creager, A.~Goldenberg, and D.~K. Duvenaud, ``Explaining image
  classifiers by counterfactual generation,'' in \emph{ICLR}, 2019.

\bibitem{uzunova_interpretable_2019}
H.~Uzunova, J.~Ehrhardt, T.~Kepp, and H.~Handels, ``Interpretable explanations
  of black box classifiers applied on medical images by meaningful
  perturbations using variational autoencoders,'' in \emph{Medical Imaging:
  Image Processing}, 2019.

\bibitem{major_interpreting_2020}
D.~Major, D.~Lenis, M.~Wimmer, G.~Sluiter, A.~Berg, and K.~B{\"u}hler,
  ``Interpreting medical image classifiers by optimization based counterfactual
  impact analysis,'' in \emph{ISBI}, 2020, pp. 1096--1100.

\bibitem{Fong2019UnderstandingDN}
R.~Fong, M.~Patrick, and A.~Vedaldi, ``Understanding deep networks via extremal
  perturbations and smooth masks,'' in \emph{ICCV}, 2019, pp. 2950--2958.

\bibitem{Wagner2019InterpretableAF}
J.~Wagner, J.~M. K{\"o}hler, T.~Gindele, L.~Hetzel, J.~T. Wiedemer, and
  S.~Behnke, ``Interpretable and fine-grained visual explanations for
  convolutional neural networks,'' in \emph{CVPR}, 2019, pp. 9089--9099.

\bibitem{dhurandhar_explanations_2018}
A.~Dhurandhar, P.-Y. Chen, R.~Luss, C.-C. Tu, P.-S. Ting, K.~Shanmugam, and
  P.~Das, ``Explanations based on the missing: Towards contrastive explanations
  with pertinent negatives,'' in \emph{NeurIPS}, 2018.

\bibitem{hsieh2020evaluations}
C.-Y. Hsieh, C.-K. Yeh, X.~Liu, P.~Ravikumar, S.~Kim, S.~Kumar, and C.-J.
  Hsieh, ``Evaluations and methods for explanation through robustness
  analysis,'' \emph{ArXiv}, vol. abs/2006.00442, 2020.

\bibitem{khakzar_explaining_2019}
A.~Khakzar, S.~Baselizadeh, S.~Khanduja, S.~T. Kim, and N.~Navab, ``Explaining
  neural networks via perturbing important learned features,'' \emph{ArXiv},
  vol. abs/1911.11081, 2019.

\bibitem{woods_adversarial_2019}
W.~Woods, J.~Chen, and C.~Teuscher, ``Adversarial explanations for
  understanding image classification decisions and improved neural network
  robustness,'' in \emph{Nature Machine Intelligence}, vol.~1, 2019, pp.
  508--516.

\bibitem{DBLP:journals/corr/abs-1807-07784}
G.~Maicas, G.~Snaauw, A.~P. Bradley, I.~D. Reid, and G.~Carneiro, ``Model
  agnostic saliency for weakly supervised lesion detection from breast
  dce-mri,'' in \emph{ISBI}, 2019, pp. 1057--1060.

\bibitem{Arjovsky2017WassersteinGA}
M.~Arjovsky, S.~Chintala, and L.~Bottou, ``Wasserstein generative adversarial
  networks,'' in \emph{ICML}, 2017.

\bibitem{DBLP:journals/corr/WangPLLBS17}
X.~Wang, Y.~Peng, L.~Lu, Z.~Lu, M.~Bagheri, and R.~M. Summers, ``Chestx-ray8:
  Hospital-scale chest x-ray database and benchmarks on weakly-supervised
  classification and localization of common thorax diseases,'' in \emph{CVPR},
  2017, pp. 3462--3471.

\bibitem{DBLP:journals/corr/HeZR016}
K.~He, X.~Zhang, S.~Ren, and J.~Sun, ``Identity mappings in deep residual
  networks,'' in \emph{ECCV}, 2016.

\bibitem{imagenet_cvpr09}
J.~Deng, W.~Dong, R.~Socher, L.-J. Li, K.~Li, and L.~Fei-Fei, ``{ImageNet: A
  Large-Scale Hierarchical Image Database},'' in \emph{CVPR}, 2009.

\bibitem{DBLP:journals/corr/KingmaB14}
D.~P. Kingma and J.~Ba, ``Adam: {A} method for stochastic optimization,'' in
  \emph{ICLR}, 2015.

\bibitem{pix2pix2017}
P.~Isola, J.-Y. Zhu, T.~Zhou, and A.~A. Efros, ``Image-to-image translation
  with conditional adversarial networks,'' in \emph{CVPR}, 2017.

\bibitem{DBLP:journals/corr/RonnebergerFB15}
O.~Ronneberger, P.~Fischer, and T.~Brox, ``U-net: Convolutional networks for
  biomedical image segmentation,'' in \emph{MICCAI}, 2015.

\bibitem{fu2020distributionguided}
W.~Fu, M.~Wang, M.~Du, N.~Liu, S.~Hao, and X.~Hu, ``Distribution-guided local
  explanation for black-box classifiers,'' 2019.

\end{thebibliography}

\end{document}